\documentclass{article}

\usepackage{microtype}
\usepackage{graphicx}
\usepackage{booktabs}
\usepackage{hyperref}

\usepackage[preprint]{icml2026}

\usepackage{amsmath}
\usepackage{amssymb}
\usepackage{mathtools}

\usepackage[capitalize,noabbrev]{cleveref}

\usepackage{algorithm}
\usepackage{algorithmic}

\icmltitlerunning{STEM Agent}

\begin{document}

\twocolumn[
  \icmltitle{STEM Agent: A Self-Adapting, Tool-Enabled, Extensible Architecture\\for Multi-Protocol AI Agent Systems}

  \begin{icmlauthorlist}
    \icmlauthor{Alfred Shen}{}
    \icmlauthor{Aaron Shen}{}
  \end{icmlauthorlist}

  \icmlcorrespondingauthor{Alfred Shen}{alfreshe@amazon.com}
  \icmlcorrespondingauthor{Aaron Shen}{aaron.shen@berkeley.edu}

  \icmlkeywords{AI Agents, Multi-Protocol, Self-Adaptation, Tool Use, Model Context Protocol}

  \vskip 0.3in
]

\printAffiliationsAndNotice{}

\begin{abstract}
Current AI agent frameworks commit early to a single interaction protocol, a fixed tool integration strategy, and static user models, limiting their deployment across diverse interaction paradigms.
We introduce \textbf{STEM Agent} (Self-adapting, Tool-enabled, Extensible, Multi-agent), a modular architecture inspired by biological pluripotency: an undifferentiated agent core that differentiates into specialized protocol handlers, tool bindings, and memory subsystems, composing into a functioning AI system.
The framework unifies five interoperability protocols (A2A, AG-UI, A2UI, UCP, and AP2) behind a single gateway, introduces a Caller Profiler that continuously learns user preferences across 20+ behavioral dimensions, externalizes all domain capabilities through the Model Context Protocol (MCP), and implements a biologically-inspired skills acquisition system where recurring interaction patterns \emph{crystallize} into reusable agent skills through a maturation lifecycle analogous to cell differentiation.
The memory system incorporates consolidation mechanisms (episodic pruning, semantic deduplication, and pattern extraction) designed for sub-linear growth under sustained interaction. A 413-test suite validates protocol handler behavior and component integration across all five architectural layers, completing in under 3\,s.
\end{abstract}

\section{Introduction}
\label{sec:intro}

In developmental biology, a stem cell is remarkable for its \emph{pluripotency}: it is undifferentiated yet capable of specializing into any cell type, which then compose into organs that sustain a living body. We take this as a design principle for AI agent systems. The \textbf{STEM Agent} (Self-adapting, Tool-enabled, Extensible, Multi-agent) is an undifferentiated agent core that \emph{differentiates} into specialized protocol handlers, tool bindings, and memory types, which \emph{compose} into a functioning system that supports diverse business workflows.

This analogy is not merely rhetorical. Current agent frameworks exhibit what we term \emph{architectural lock-in}: they commit early to a single interaction protocol (e.g., REST-only or chat-only), a fixed tool integration strategy, and static user models. The result is an ecosystem of rigid, single-purpose agents that cannot interoperate, adapt, or compose~\citep{ferrag2025agents, tran2025multiagent}. Meanwhile, the proliferation of agent communication protocols---MCP~\citep{ehtesham2025survey}, A2A~\citep{habler2025a2a}, and emerging standards for UI streaming and commerce---demands architectures that are protocol-pluralistic by design~\citep{li2025gluetoprotocols}.

STEM Agent addresses these gaps with the following contributions:

\begin{enumerate}
\item \textbf{Multi-protocol interoperability.} To our knowledge, the first agent framework implementing five interoperability protocols---A2A (agent-to-agent), AG-UI (streaming UI events), A2UI (dynamic UI composition), UCP (universal commerce), and AP2 (agent payments)---behind a unified gateway. Of these, A2A and AG-UI follow published external specifications; UCP and AP2 are novel protocols proposed herein.

\item \textbf{Caller Profiler.} A multi-dimensional user modeling system that continuously learns caller preferences across 4 categories and 20+ dimensions using exponential moving averages, enabling per-user behavioral adaptation without manual configuration.

\item \textbf{MCP-native tool integration.} All external \emph{domain} capabilities are acquired at runtime via the Model Context Protocol, separating agent reasoning from domain knowledge. Meta-reasoning rules (strategy selection, parameter tuning) remain in code.

\item \textbf{Self-tunable behavior parameters.} Ten continuously adjusted parameters (reasoning depth, creativity, verbosity, etc.) that adapt to task characteristics and caller profiles.

\item \textbf{Commerce-ready agent protocols.} Novel UCP and AP2 protocol handlers enabling checkout sessions, mandate-based payments, and audit trails within the agent interaction loop.

\item \textbf{Biologically-inspired skills acquisition.} A cell-differentiation model where recurring interaction patterns crystallize into reusable skills that mature through activation (progenitor $\to$ committed $\to$ mature) or undergo apoptosis on persistent failure, complemented by manual skill plugin support.
\end{enumerate}

\section{Related Work}
\label{sec:related}

\paragraph{Multi-agent frameworks.}
The rapid growth of LLM-based agent systems has produced a diverse ecosystem of frameworks, each optimizing for different coordination patterns. \citet{tran2025multiagent} survey these mechanisms---role-assignment, debate, and orchestration---across AutoGen, MetaGPT, CAMEL, and CrewAI. A recurring finding is that most frameworks commit to a single communication protocol and lack per-user adaptation. \citet{orogat2026mafbench} confirm this in MAFBench, a unified benchmark showing that architectural choices drive 100$\times$ latency differences and 30\% accuracy gaps across frameworks. \citet{dang2025evolving} address coordination rigidity through evolving orchestration, while \citet{drammeh2025orchestration} demonstrate that multi-agent LLM orchestration can achieve deterministic decision support for incident response. STEM Agent builds on this line of work by unifying five protocols behind a single gateway and adding continuous caller modeling.

\paragraph{Agent communication protocols.}
The interoperability landscape is fragmented across competing standards. \citet{ehtesham2025survey} survey four major protocols---MCP, ACP, A2A, and ANP---concluding that MCP and A2A are complementary (vertical tool access vs.\ horizontal agent communication) rather than competing. \citet{li2025gluetoprotocols} critically analyze integration challenges when combining A2A and MCP, identifying schema translation and lifecycle management as key pain points. \citet{jeong2025mcpa2a} study the MCP$\times$A2A framework. Security is an active concern: \citet{habler2025a2a} analyze A2A threat models, while \citet{anbiaee2026security} provide a comparative security analysis across MCP, A2A, Agora, and ANP. \citet{sarkar2025mcpsurvey} map classical design patterns (Mediator, Observer) to MCP communication. The Agent Network Protocol~\citep{chang2025anp} addresses decentralized discovery. Despite this proliferation, prior work implements at most two protocols; STEM Agent implements five, including two novel commerce protocols.

\paragraph{Agentic reasoning and adaptive compute.}
\citet{wei2026agentic} provide a comprehensive survey of agentic reasoning, covering chain-of-thought, ReAct, reflexion, tree-of-thought, and debate paradigms. \citet{alomrani2025reasoning} survey adaptive test-time compute, showing that dynamically adjusting reasoning depth yields efficiency gains without quality loss. \citet{li2025budget} propose budget-guided steering for LLM thinking. STEM Agent integrates four reasoning strategies with automatic selection based on task characteristics, drawing on these adaptive compute principles.

\paragraph{MCP tools and benchmarks.}
\citet{luo2025mcpuniverse} benchmark LLMs across 11 real-world MCP servers where even state-of-the-art models score below 44\%; \citet{fan2025mcptoolbench} scale this to 4,000+ servers. \citet{lumer2025scalemcp} propose dynamic tool synchronization, and \citet{hasan2026mcpsmells} find that 97.1\% of MCP tool descriptions contain quality issues. \citet{jayanti2026enhancing} and \citet{schlapbach2026convergence} extend MCP with context-aware collaboration and schema-guided dialogue convergence, respectively.

\paragraph{Agent memory and self-adaptation.}
\citet{ferrag2025agents} review the progression from LLM reasoning to autonomous AI agents, identifying continuous adaptation as a key open challenge. \citet{ai2025memorybench} benchmark memory and continual learning in LLM systems. \citet{gallego2026memoryastool} propose distilling feedback into memory-as-a-tool, treating memory as an explicit, learnable capability. \citet{yadav2026synapse} present a hierarchical multi-agent framework with hybrid memory. STEM Agent operationalizes these ideas through a four-type memory system and a Caller Profiler with self-tunable behavior parameters.

\section{Architecture}
\label{sec:architecture}

\begin{figure*}[!ht]
  \begin{center}
    \includegraphics[width=\textwidth]{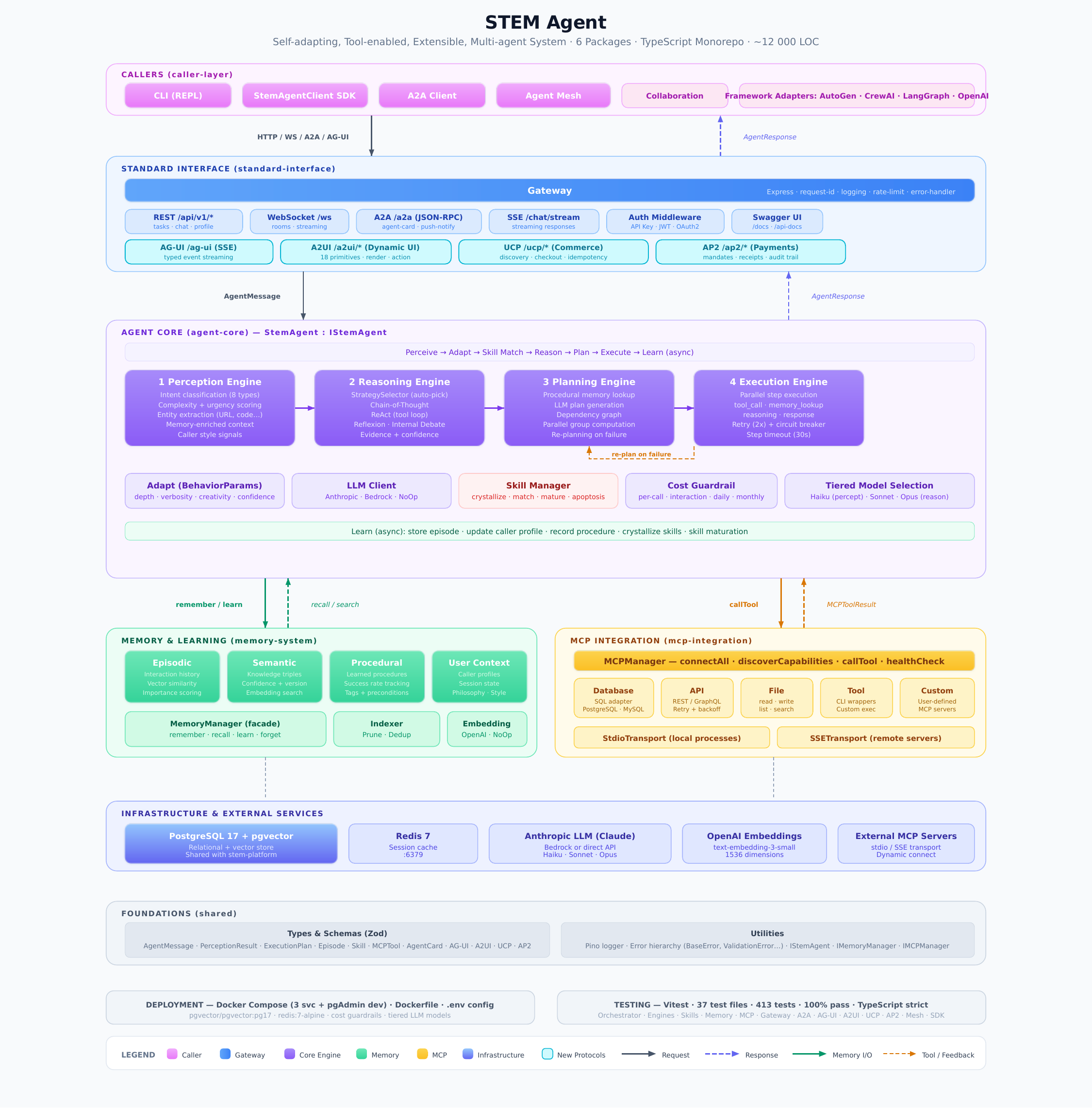}
  \end{center}
  \caption{STEM Agent five-layer architecture. Callers interact through the Standard Interface Layer, which routes requests through five protocol handlers (A2A, AG-UI, A2UI, UCP, AP2) and framework adapters to the Agent Core. The core's cognitive pipeline (Perceive $\to$ Adapt $\to$ Skill Match $\to$ Reason $\to$ Plan $\to$ Execute $\to$ Learn $\to$ Respond) is supported by a four-type Memory System and the MCP Integration Layer for dynamic tool access.}
  \label{fig:architecture}
\end{figure*}

STEM Agent is organized into five layers, each corresponding to a distinct concern (\cref{tab:layers}). \Cref{fig:architecture} provides a visual overview. Following the stem cell analogy: Layer~3 (Agent Core) is the undifferentiated core that differentiates through Layer~2 (protocol handlers) and Layer~5 (tool bindings) into specialized capabilities, while Layer~4 (Memory) provides persistent state that guides future adaptation.

\begin{table}[t]
  \caption{Five-layer architecture overview.}
  \label{tab:layers}
  \begin{center}
    \begin{small}
      \begin{tabular}{clc}
        \toprule
        Layer & Component & Key Counts \\
        \midrule
        1 & Caller / User Layer & 4 adapters \\
        2 & Standard Interface (Gateway) & 5 protocols \\
        3 & Agent Core & 5 engines \\
        3.5 & Security Layer & 4 IAM plugins \\
        4 & Memory System & 4 memory types \\
        5 & MCP Integration & 3 transports \\
        \bottomrule
      \end{tabular}
    \end{small}
  \end{center}
  \vskip -0.1in
\end{table}

\subsection{Cognitive Pipeline}

The Agent Core processes each request through an eight-phase cognitive pipeline (\cref{alg:pipeline}).

\begin{algorithm}[t]
  \caption{STEM Cognitive Pipeline}
  \label{alg:pipeline}
  \begin{algorithmic}
    \STATE {\bfseries Input:} message $m$, caller context $c$
    \STATE {\bfseries Output:} response $r$, updated profile $c'$
    \STATE
    \STATE \textit{// Phase 1: Perception}
    \STATE $p \leftarrow \textsc{Perceive}(m)$ \hfill $\triangleright$ intent, entities, complexity
    \STATE
    \STATE \textit{// Phase 2: Adaptation}
    \STATE $\theta \leftarrow \textsc{Adapt}(c, p)$ \hfill $\triangleright$ load profile, adjust params
    \STATE
    \STATE \textit{// Phase 3: Skill Match}
    \STATE $\sigma \leftarrow \textsc{MatchSkills}(p)$ \hfill $\triangleright$ check acquired skills
    \IF{$\sigma \neq \emptyset$ \AND $\sigma_1.\text{maturity} \geq \text{committed}$}
      \STATE $\mathcal{P} \leftarrow \textsc{SkillToPlan}(\sigma_1)$ \hfill $\triangleright$ short-circuit
    \ELSE
      \STATE \textit{// Phase 4: Reasoning}
      \STATE $s \leftarrow \textsc{SelectStrategy}(p, \theta)$
      \STATE $\mathcal{R} \leftarrow \textsc{Reason}(m, p, \theta, s)$
      \STATE \textit{// Phase 5: Planning}
      \STATE $\mathcal{P} \leftarrow \textsc{Plan}(\mathcal{R}, \textsc{McpTools}())$
    \ENDIF
    \STATE
    \STATE \textit{// Phase 6: Execution}
    \STATE $\mathcal{E} \leftarrow \textsc{Execute}(\mathcal{P})$ \hfill $\triangleright$ MCP tool calls
    \STATE
    \STATE \textit{// Phase 7: Formatting}
    \STATE $r \leftarrow \textsc{Format}(\mathcal{E}, \theta)$ \hfill $\triangleright$ style for caller
    \STATE
    \STATE \textit{// Phase 8: Learning (async)}
    \STATE $c' \leftarrow \textsc{UpdateProfile}(c, p)$
    \STATE $\textsc{RecordSkillOutcome}(\sigma_1, \mathcal{E}.\text{success})$
    \STATE $\textsc{TryCrystallize}()$ \hfill $\triangleright$ pattern $\to$ new skill
    \STATE \textbf{return} $r, c'$
  \end{algorithmic}
\end{algorithm}

The \textsc{Perceive} phase classifies intent into 10 categories, estimates complexity as one of three levels (\emph{simple}, \emph{medium}, \emph{complex}) based on word count, entity density, and code presence, and extracts entities, sentiment, and urgency. The \textsc{Adapt} phase loads the caller's learned profile and adjusts behavior parameters accordingly. The \textsc{MatchSkills} phase checks the skill registry for acquired skills matching the current perception; if a committed or mature skill matches, its pre-built plan is used directly, bypassing \textsc{Reason} and \textsc{Plan} (see \cref{sec:skills} for the full skills lifecycle). Otherwise, \textsc{SelectStrategy} chooses among four reasoning strategies based on task characteristics: tool-requiring tasks use ReAct, complex tasks trigger Reflexion, analysis and creative requests use Internal Debate, and all others default to Chain-of-Thought (see \cref{sec:strategy} for details). The \textsc{Plan} phase selects MCP tools and constructs an execution plan with parallel steps where dependencies allow. \textsc{Execute} orchestrates MCP tool calls with retries (default: 2) and a circuit breaker (threshold: 3 consecutive failures). The \textsc{Learning} phase runs asynchronously: it updates the caller profile, records skill activation outcomes (advancing maturity or triggering apoptosis), and attempts to crystallize new skills from accumulated episode patterns.

\subsection{Implementation}

The system is implemented as a TypeScript monorepo with six workspace packages: \texttt{shared} (154+ Zod schemas for runtime type safety), \texttt{agent-core} (cognitive engines), \texttt{standard-interface} (protocol handlers and gateway), \texttt{mcp-integration} (MCP client layer), \texttt{memory-system} (four memory stores), and \texttt{caller-layer} (caller utilities). The gateway is built on Express.js~5, with each protocol handler mounted via a pluggable \texttt{createRouter()} pattern.

\section{Multi-Protocol Interoperability}
\label{sec:protocols}

A key contribution of STEM Agent is the simultaneous support for five interaction protocols behind a unified gateway. \Cref{tab:protocols} summarizes each protocol.

\begin{table*}[t]
  \caption{Protocol comparison. STEM Agent implements all five behind a single Express.js gateway. A2A follows the Linux Foundation v0.3.0 specification. UCP and AP2 are novel protocols proposed in this work.}
  \label{tab:protocols}
  \begin{center}
    \begin{small}
      \begin{tabular}{llllp{5.5cm}}
        \toprule
        Protocol & Endpoint & Transport & Schemas & Use Case \\
        \midrule
        A2A & \texttt{POST /a2a} & JSON-RPC 2.0 & 6 & Agent-to-agent task delegation, inter-agent messaging, agent card discovery \\
        AG-UI & \texttt{POST /ag-ui} & SSE & 20 & Real-time UI streaming; maps pipeline phases to typed events \\
        A2UI & \texttt{POST /a2ui/render} & SSE + REST & 19 & Dynamic UI composition with 16 component primitives \\
        UCP & \texttt{POST /ucp/*} & REST & 14 & Commerce checkout sessions with idempotency \\
        AP2 & \texttt{POST /ap2/*} & REST & 10 & Mandate-based payments with audit trail \\
        \bottomrule
      \end{tabular}
    \end{small}
  \end{center}
  \vskip -0.1in
\end{table*}

\paragraph{A2A (Agent-to-Agent).} Implements the A2A v0.3.0 specification (Linux Foundation) using JSON-RPC 2.0. Supports \texttt{tasks/send}, \texttt{tasks/sendSubscribe} (SSE streaming), \texttt{tasks/get}, and \texttt{tasks/cancel}. Agent discovery is served at \texttt{/.well-known/agent.json}.

\paragraph{AG-UI (Agent-User Interaction).} Streams cognitive pipeline events to frontends via Server-Sent Events. Events include \texttt{TEXT\_MESSAGE\_START}, \texttt{REASONING\_MESSAGE}, \texttt{TOOL\_CALL\_START}, \texttt{STATE\_SNAPSHOT}, and \texttt{RUN\_FINISHED}, enabling fine-grained progress rendering.

\paragraph{A2UI (Agent-to-User Interface).} Provides dynamic UI composition through 16 component primitives organized in a flat adjacency list model where each component references children by ID, enabling flexible layout construction without a fixed widget hierarchy.

\paragraph{UCP (Universal Commerce Protocol).} Manages checkout session lifecycles with required idempotency headers (\texttt{Idempotency-Key}, \texttt{Request-Id}, \texttt{UCP-Agent}). Sessions progress through creation, retrieval, and completion endpoints. An idempotency cache prevents duplicate session creation.

\paragraph{AP2 (Agent Payments Protocol).} Implements a three-phase payment lifecycle: Intent Mandate $\rightarrow$ Payment Mandate $\rightarrow$ Payment Receipt. Supports auto-approval for payments below a configurable threshold and maintains a full audit trail per intent.

\paragraph{Gateway architecture.} Each protocol handler implements a \texttt{createRouter(): Router} method returning an Express.js router. The gateway mounts all routers with shared middleware for authentication, rate limiting, request correlation, and error handling. Adding a new protocol requires only implementing the handler and mounting its router. Four framework adapters (AutoGen, CrewAI, LangGraph, OpenAI Agents SDK) translate external conventions to STEM Agent's internal format.

\section{Self-Adaptation and Learning}
\label{sec:adaptation}

STEM Agent's self-adaptation operates at two levels: per-caller profile learning and per-task behavior parameter tuning.

\subsection{Caller Profiler}

The Caller Profiler learns a multi-dimensional model of each user across four categories (\cref{tab:profile}): philosophy (8 dimensions, e.g., pragmatism vs.\ idealism, risk tolerance), principles (4 dimensions, e.g., correctness over speed, testing emphasis), style (5 dimensions, e.g., formality, verbosity, technical depth), and habits (temporal and behavioral patterns).

\begin{table}[t]
  \caption{Caller Profile dimensions by category.}
  \label{tab:profile}
  \begin{center}
    \begin{small}
      \begin{tabular}{lcp{3.4cm}}
        \toprule
        Category & Dims & Example Dimensions \\
        \midrule
        Philosophy & 8 & pragmatism vs.\ idealism, risk tolerance, innovation orientation \\
        Principles & 4 & correctness over speed, testing emphasis, security mindedness \\
        Style & 5 & formality, verbosity, technical depth, structure preference \\
        Habits & 4+ & session length, iteration tendency, peak hours \\
        \bottomrule
      \end{tabular}
    \end{small}
  \end{center}
  \vskip -0.1in
\end{table}

\paragraph{Learning mechanism.}
Each dimension is updated via an exponential moving average (EMA) with learning rate $\alpha = 0.1$:
\begin{equation}
  v_{t+1} = (1 - \alpha) \cdot v_t + \alpha \cdot s_t
\label{eq:ema}
\end{equation}
where $v_t$ is the current profile value and $s_t$ is the signal extracted from the current interaction. We chose $\alpha = 0.1$ to balance responsiveness to new signals against stability of learned preferences; smaller values (e.g., 0.05) produced sluggish adaptation in preliminary testing, while larger values (e.g., 0.3) caused oscillation.

\paragraph{Confidence-gated adaptation.}
Profile confidence follows a rational saturation curve:
\begin{equation}
  \text{conf}(n) = \frac{n}{n + \kappa}
\label{eq:confidence}
\end{equation}
where $n$ is the number of interactions and $\kappa = 10$ is a half-life constant (conf reaches 0.5 at $n = \kappa$). Below $n = 5$ (conf $\approx 0.33$), the system relies primarily on signals from the current message; as confidence grows, the learned profile is blended with current signals weighted by confidence.

\subsection{Behavior Parameters}

Ten parameters are continuously adjusted based on task characteristics and caller profile: \emph{reasoning depth} (default: 3), \emph{exploration vs.\ exploitation} (0.3), \emph{verbosity} (0.5), \emph{confidence threshold} (0.7), \emph{tool use preference} (0.5), \emph{creativity} (0.5), \emph{proactive suggestion} (on), \emph{self-reflection frequency} (every 5 steps), \emph{max plan steps} (10), and \emph{memory retrieval breadth} (10). The Adaptation phase adjusts these using the caller profile and current task perception as inputs.

\subsection{Memory System}

The memory system implements four complementary types informed by recent agent memory research~\citep{ai2025memorybench, gallego2026memoryastool}:

\begin{enumerate}
\item \textbf{Episodic memory}: Stores specific interaction episodes with vector embeddings for similarity search (PostgreSQL + pgvector). Each episode carries an importance score.
\item \textbf{Semantic memory}: Maintains knowledge triples (subject, predicate, object) in concept graphs, with patterns extracted from episodic memory.
\item \textbf{Procedural memory}: Records successful strategies and tool usage patterns, enabling best-procedure matching for recurring task types.
\item \textbf{User context memory}: Per-caller session history and profiles with GDPR forget-me support.
\end{enumerate}

A Memory Manager provides a unified facade, delegating to specialized modules and performing memory consolidation (episodic $\rightarrow$ semantic/procedural) as interaction history grows.

\subsection{Reasoning Strategy Selection}
\label{sec:strategy}

The Strategy Selector maps task characteristics to reasoning strategies via deterministic rules:

\begin{itemize}
\item Tool-requiring tasks $\rightarrow$ \textbf{ReAct}~\citep{wei2026agentic}
\item Complexity = \emph{complex} $\rightarrow$ \textbf{Reflexion}~\citep{wei2026agentic}
\item Analysis or creative requests $\rightarrow$ \textbf{Internal Debate}~\citep{tran2025multiagent}
\item Default $\rightarrow$ \textbf{Chain-of-Thought}~\citep{wei2026agentic}
\end{itemize}

These rules are deliberately simple; the classification of task complexity and intent (computed during the Perception phase) provides the necessary signal. We note that this selection logic is \emph{not} externalized via MCP---it is meta-reasoning that we consider part of the agent core rather than domain knowledge.

\subsection{Skills Acquisition via Cell Differentiation}
\label{sec:skills}

Extending the stem cell metaphor, we model skill acquisition as cell differentiation: internal cues (episodic memory patterns, procedure success rates) and external cues (MCP tool availability, caller domain signals) trigger \emph{crystallization} of specialized skills. A skill encapsulates a trigger condition (intent patterns, domains, entity types), an action sequence (tool chain or procedure steps), and maturity metadata. Skills progress through three stages mirroring cell lineage:

\begin{enumerate}
\item \textbf{Progenitor}: newly crystallized from episode patterns; not yet used for shortcutting.
\item \textbf{Committed}: after $k_c{=}3$ successful activations with success rate ${\geq}0.6$; can short-circuit the Reason${\to}$Plan pipeline.
\item \textbf{Mature}: after $k_m{=}10$ successful activations; receives matching priority.
\end{enumerate}

\paragraph{Crystallization and apoptosis.}
The Learn phase groups recent episodes by action patterns; when ${\geq}3$ episodes share a common action key and topic keywords appear in ${\geq}50\%$ of episodes, a progenitor skill is created. Conversely, crystallized skills with success rate ${<}0.3$ after ${\geq}10$ activations are removed (\emph{apoptosis}). Users can also manually register or remove \emph{plugin skills} (induced differentiation), bypassing the crystallization process.

\section{Evaluation}
\label{sec:evaluation}

\subsection{Test Suite and Protocol Compliance}

The test suite comprises 413 tests across 37 test files using Vitest, with 100\% pass rate and a total runtime of 2.92\,s. Tests cover unit tests per engine (perception, reasoning, planning, execution, skills acquisition), protocol handler integration tests (verifying A2A JSON-RPC compliance, AG-UI event sequences, UCP idempotency, AP2 audit trails), memory system tests, MCP integration tests, security middleware tests, framework adapter tests, and gateway end-to-end tests. While test count alone does not guarantee correctness, the breadth of coverage across all five layers provides confidence in cross-component integration.

\subsection{Architectural Overhead Analysis}

Each protocol handler adds a thin routing and serialization layer atop the shared cognitive pipeline. A2A requires JSON-RPC envelope parsing and task state management. AG-UI requires only SSE channel setup. A2UI adds component tree construction and layout validation. UCP adds idempotency key lookup and checkout state management. AP2 adds mandate validation and audit trail writes. Since all handlers delegate to the same Agent Core pipeline with mocked LLM inference, protocol-specific overhead is dominated by serialization and middleware traversal rather than computation. Formal latency benchmarking under controlled conditions is left to future work.

\subsection{Framework Comparison}

\Cref{tab:comparison} compares STEM Agent with existing frameworks across architectural dimensions (feature comparison, not performance benchmark).

\begin{table}[t]
  \caption{Architectural comparison with existing agent frameworks. Proto.\ = standardized interoperability protocols; Adapt = per-caller self-adaptive behavior; Mem.\ = distinct memory subsystem types; Skills = emergent skill acquisition; Com.\ = commerce protocols. $\dagger$UCP and AP2 are novel protocols proposed in this work.}
  \label{tab:comparison}
  \begin{center}
    \begin{small}
      \begin{tabular}{lccccc}
        \toprule
        Framework & Proto. & Adapt & Mem. & Skills & Com. \\
        \midrule
        AutoGen & 1 & \texttimes & 1 & \texttimes & \texttimes \\
        MetaGPT & 1 & \texttimes & 1 & \texttimes & \texttimes \\
        CrewAI & 1 & \texttimes & 1 & \texttimes & \texttimes \\
        LangChain & 1 & \texttimes & 2 & \texttimes & \texttimes \\
        \textbf{STEM} & \textbf{3+2$^\dagger$} & \checkmark & \textbf{4} & \checkmark & \checkmark \\
        \bottomrule
      \end{tabular}
    \end{small}
  \end{center}
  \vskip -0.1in
\end{table}

We count ``standardized protocols'' as those with published, externally-maintained specifications. STEM Agent implements three established protocols (A2A v0.3.0, AG-UI, MCP) plus two novel ones (UCP, AP2), validated through integration tests verifying event sequences, error handling, and idempotency. The pluggable IAM architecture supports four authentication plugins (JWT, OAuth2, SAML, API Key) with TTL-based policy caching and configurable rate limiting.

\section{Limitations}
\label{sec:limitations}

\paragraph{No end-to-end benchmark evaluation.} Our evaluation validates compliance through testing but does not include task-completion benchmarks (e.g., MAFBench~\citep{orogat2026mafbench}) or user studies. \textbf{Adaptation simplicity:} The EMA-based Caller Profiler cannot capture non-stationary or multi-modal preference distributions. \textbf{Commerce protocol maturity:} UCP and AP2 lack external adoption and formal threat modeling. \textbf{Scalability:} In-memory stores for profiles and idempotency caches require distributed locking under high concurrency.

\section{Conclusion and Future Work}
\label{sec:conclusion}

We have presented STEM Agent, demonstrating that protocol plurality, behavioral self-adaptation, and emergent skill acquisition can coexist when external capabilities are mediated by MCP and user modeling is decoupled from domain logic. The biologically-inspired skills system extends the stem cell metaphor to runtime behavior: the agent crystallizes reusable skills from recurring patterns, with maturation and apoptosis ensuring quality.

Future work includes: \emph{STEM Platform}, an orchestration layer that dynamically integrates and composes multiple STEM Agents into complex, multi-agent projects through delegation, consensus, and pipeline collaboration patterns; benchmark evaluation~\citep{orogat2026mafbench, luo2025mcpuniverse}; learned strategy selection via adaptive test-time compute~\citep{alomrani2025reasoning}; interoperability with ANP~\citep{chang2025anp} and AWCP~\citep{nie2026awcp}; embedding-based skill matching and cross-session skill transfer; and formal threat modeling for UCP/AP2~\citep{shen2026mcp38, anbiaee2026security}.

\section*{Impact Statement}

This paper presents work whose goal is to advance the field of AI agent systems. The STEM Agent architecture is designed for transparency (audit logging, explainable adaptation) and user control (GDPR forget-me support, configurable autonomy levels). As with any agent framework, deployment should consider safety boundaries, particularly when commerce protocols handle financial transactions. We encourage adopters to implement appropriate guardrails for their deployment context. The source code is available at \url{https://github.com/alfredcs/stem-agent} under the MIT License.

\bibliography{references}
\bibliographystyle{icml2026}

\end{document}